\documentclass[10pt,twocolumn,letterpaper]{article}

\usepackage{wacv}
\usepackage{times}
\usepackage{epsfig}
\usepackage{graphicx}
\usepackage{comment}
\usepackage{amsmath,amssymb} % define this before the line numbering.
\usepackage{booktabs}
\usepackage{algorithm}
\usepackage{algorithmic}
\usepackage{color}
\usepackage[small]{caption}

\usepackage{colortbl}
\usepackage{subcaption}

\definecolor{lightgray}{gray}{0.85}

\DeclareMathOperator*{\argmin}{arg\,min}
\newcommand{\norm}[1]{\left\lVert#1\right\rVert}

\def\wacvPaperID{198} % Enter the WACV Paper ID here

\wacvfinalcopy % *** Uncomment this line for the final submission

\ifwacvfinal
\def\assignedStartPage{1} % *** Enter the assigned starting page number (instead of 9876)
\fi

\ifwacvfinal
\else
\fi

\begin{document}

%%%%%%%%% TITLE
\title{Multi-Modal Trajectory Prediction of NBA Players}
\author{Sandro Hauri$^1$, Nemanja Djuric$^2$, Vladan Radosavljevic$^3$, Slobodan Vucetic$^1$\\
$^1$Temple University, $^2$Uber Advanced Technology Group, $^3$Spotify\\
{\tt\small sandro.hauri@temple.edu, ndjuric@uber.com, vladanr@spotify.com, vucetic@temple.edu}
}

\maketitle
%\thispagestyle{empty}

%%%%%%%%% ABSTRACT
\begin{abstract}
National Basketball Association (NBA) players are highly motivated and skilled experts that solve complex decision making problems at every time point during a game. As a step towards understanding how players make their decisions, we focus on their movement trajectories during games. We propose a method that captures the multi-modal behavior of players, where they might consider multiple trajectories and select the most advantageous one. The method is built on an LSTM-based architecture predicting multiple trajectories and their probabilities, trained by a multi-modal loss function that updates the best trajectories. Experiments on large, fine-grained NBA tracking data show that the proposed method outperforms the state-of-the-art. In addition, the results indicate that the approach generates more realistic trajectories and that it can learn individual playing styles of specific players.
\end{abstract}

%%%%%%%%% BODY TEXT
\section{Introduction}
In recent years, advances in artificial intelligence and computer vision started revolutionizing how athletic performance and results are being analyzed and understood, which includes the use of fine-grained player tracking data during sporting events. In our research we are developing new methods aimed at deeper understanding of the behavior of athletes in team sports, with particular focus on their motion prediction. This is a particularly important task in invasion sports, such as soccer or basketball, where knowledge of how and where the players will move, especially when it comes to those from the opposing team, is of critical importance for gaining tactical advantage during a game \cite{lamas2014invasion}. Beyond this use case the benefits of accurate motion prediction extend to other applications, such as postgame analysis \cite{gudmundsson2017spatio} or improving TV broadcasting of games by optimizing camera movement \cite{chen2016learning, kim2010motion}. 
Prediction of human trajectories can also be used to improve tracking accuracy \cite{kitani2012activity}, and has recently become a vibrant topic of research in the computer vision community \cite{Alahi_2016social, Felsen_2018_ECCV, haddad2020self}.

\begin{figure*}[!ht]
\centering
  \begin{subfigure}{0.32\textwidth}
  \includegraphics[width=\linewidth]{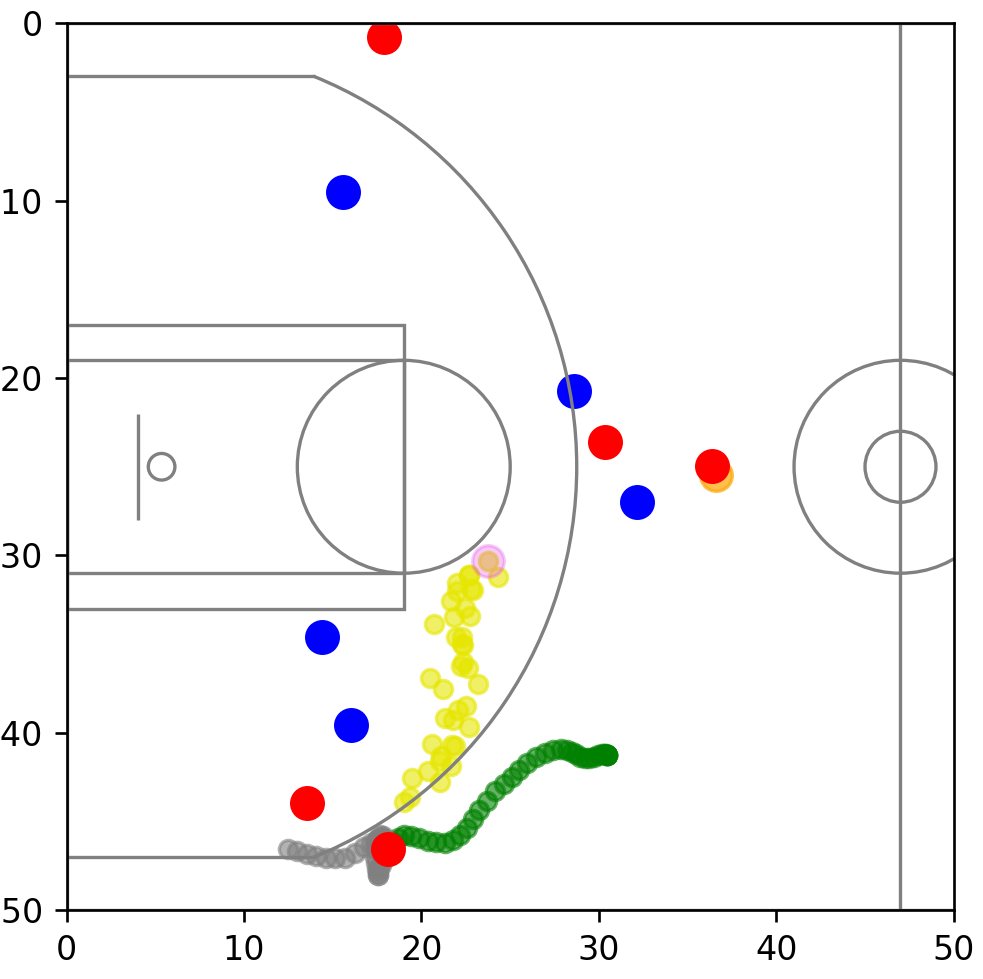}
  \vspace{-5mm}
  \caption{}
  \label{fig:path_loc}
\end{subfigure}\hfill
\begin{subfigure}{0.32\textwidth}%
  \includegraphics[width=\linewidth]{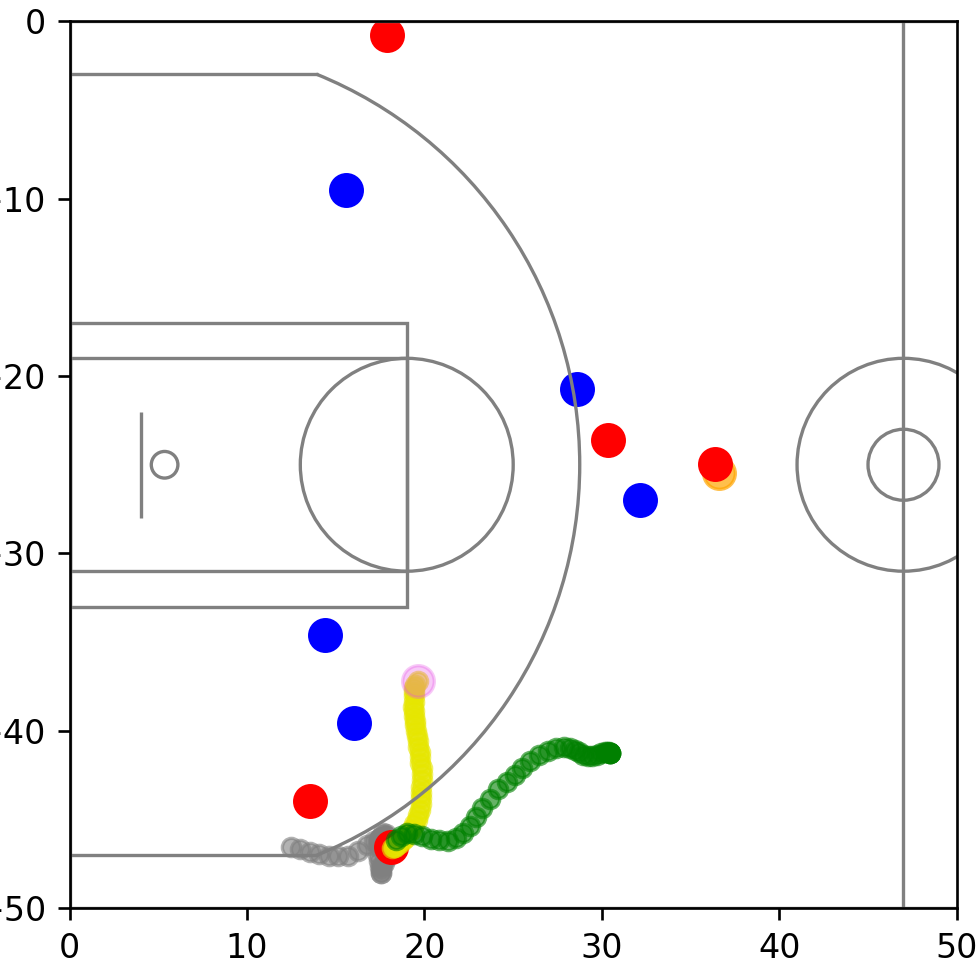}
  \vspace{-5mm}
  \caption{}
  \label{fig:path_CNN}
\end{subfigure}\hfill
\begin{subfigure}{0.32\textwidth}
  \includegraphics[width=\linewidth]{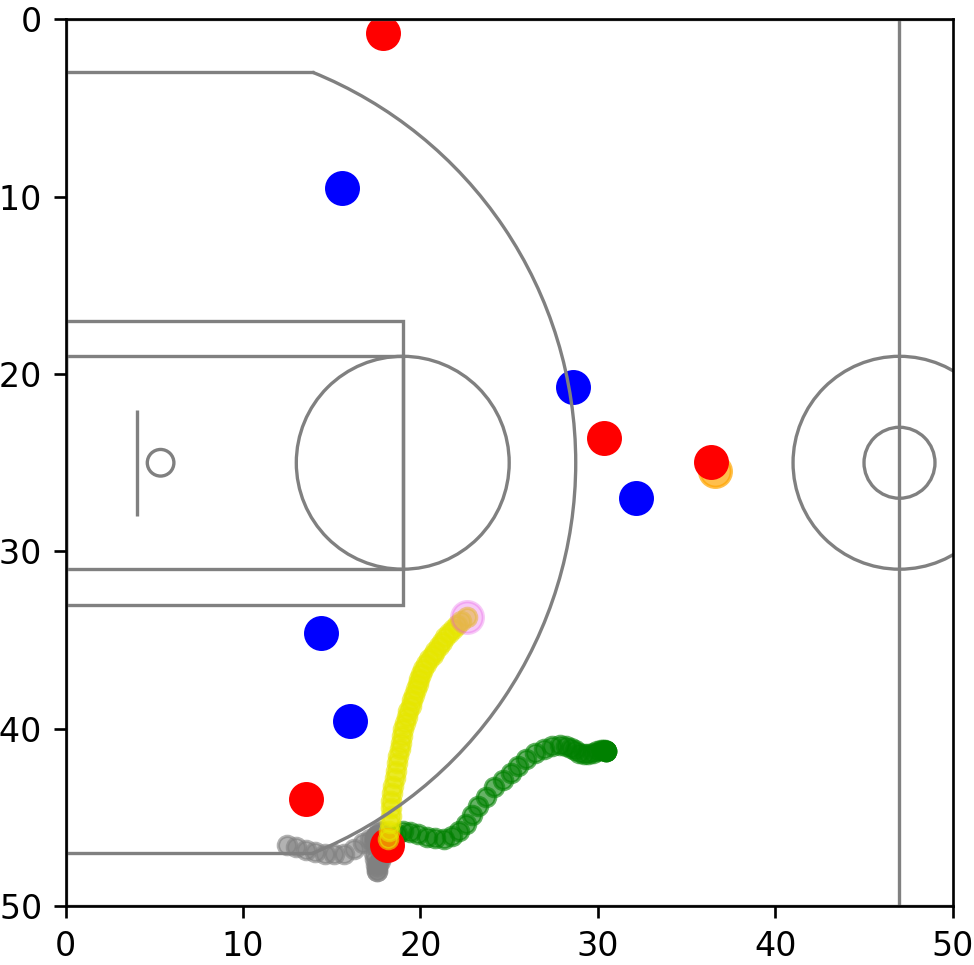}
  \vspace{-5mm}
  \caption{}
  \label{fig:path_vel}
\end{subfigure}
\newline
\begin{subfigure}{0.32\textwidth}%
  \includegraphics[width=\linewidth]{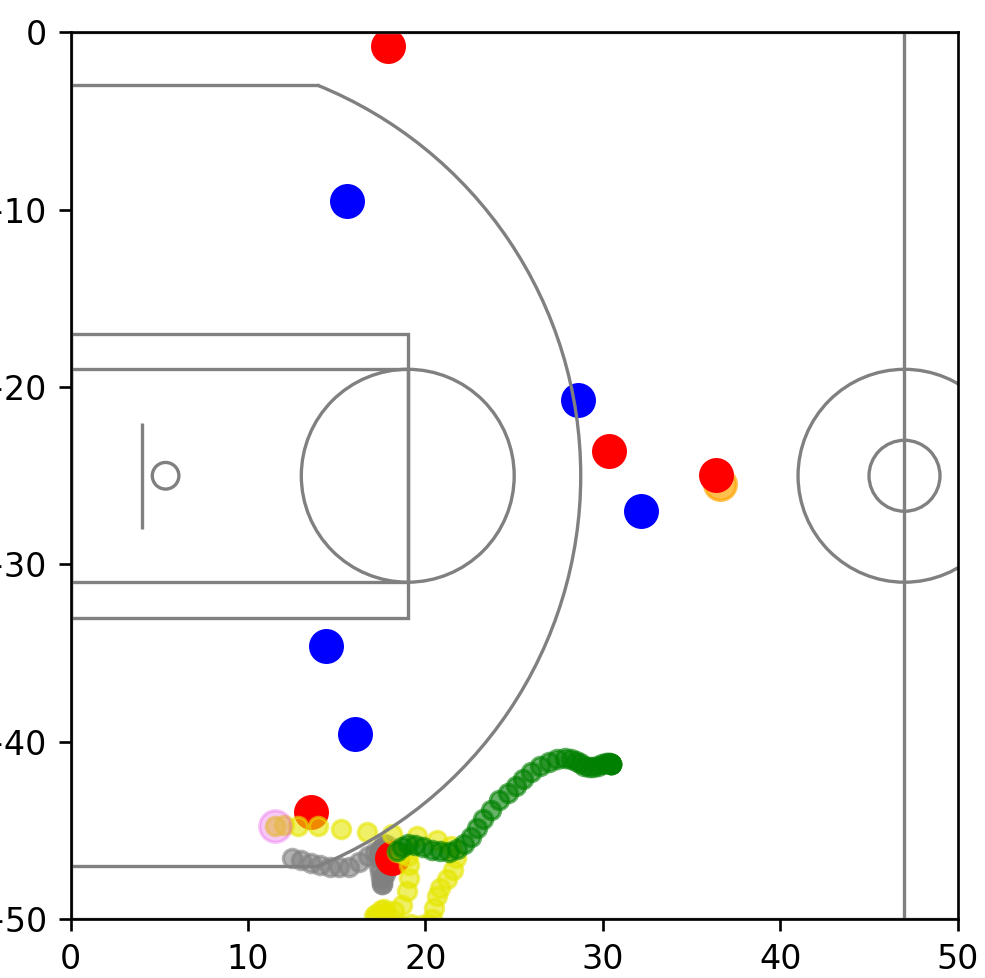}
  \vspace{-5mm}
  \caption{}
  \label{fig:path_VRNN}
\end{subfigure}\hfill
\begin{subfigure}{0.32\textwidth}%
  \includegraphics[width=\linewidth]{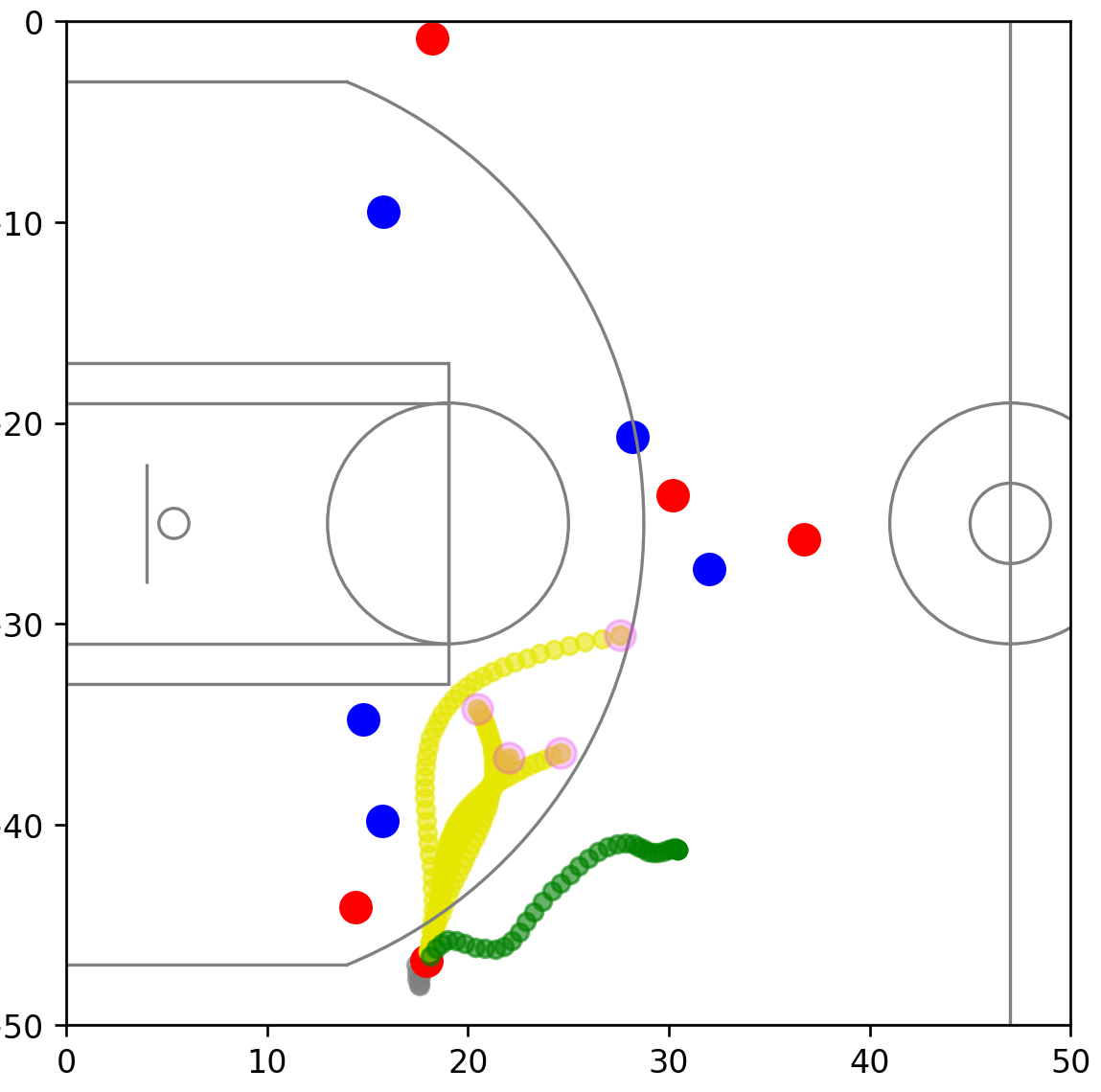}
  \vspace{-5mm}
  \caption{}
  \label{fig:path_SGAN}
\end{subfigure}\hfill
\begin{subfigure}{0.32\textwidth}%
  \includegraphics[width=\linewidth]{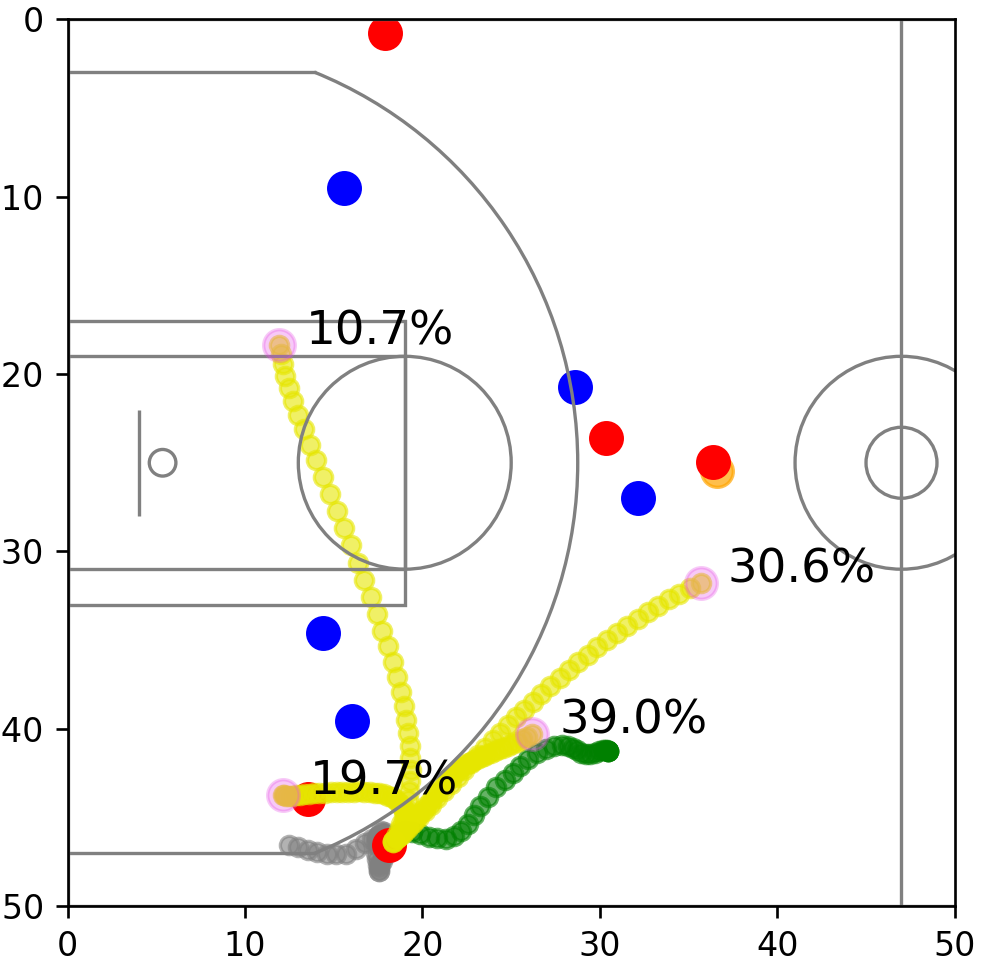}
  \vspace{-5mm}
  \caption{}
  \label{fig:path_multi}
\end{subfigure}
  \vspace{-2.5mm}
  \caption[caption]{Visualization of predicted trajectories with $H=40$ using several state-of-the-art methods: a) location-LSTM; b) CNN; c) MBT$_1$; d) MACRO VRNN$_1$, e) SocialGAN$_4$; f) MBT$_{4l}$ (ours); red: attackers, blue: defenders, orange: ball, grey: input history of predicted player, yellow: prediction, green: ground truth. A video animation is included in the Supplementary Material.}
  %\vspace{-3mm}
  \label{fig:paths}
\end{figure*}

Using mathematics, statistics, and artificial intelligence to analyze sports performance is however not a novel idea. It has been famously explored in baseball \cite{lewis2004moneyball} and applied with great success to soccer \cite{kuper2014soccernomics}, with authors uncovering useful patterns in the sports data that can be and have been used to move the needle in this highly competitive field. 
%As these advanced tools have been proven to help the clubs gain valuable insights and improve their on-field performance, the practice of statistical analysis has been adopted by many top-performing teams regardless of the sport they play.
Today, elite teams from across the globe, such as Golden State Warriors, New York Yankees, and Manchester United, have analytics departments focusing on deriving knowledge from large amounts of data these teams generate. Beyond the sports professionals, it is interesting that even general public is becoming more accepting of these complex statistical tools, as exemplified by the introduction of the concept of expected goals \cite{pollard2004xgoals} in some postgame summaries in the Premier League, the English top soccer division.
This trend is also exemplified by a number of research publications, as well as high-profile conferences and workshops organized on the topic, such as MIT Sloan SAC or KDD Sports Analytics \cite{brefeld2017guest}. These are attended by both the scientific community on one side and world-class athletes and management of professional sports teams on the other, indicating the value and benefits that the artificial intelligence is bringing to this multi-billion dollar industry.

In this paper we focus on movement prediction of NBA players during offensive possessions. Players at any moment have freedom to consider several options for their movement. Potential trajectories depend on the state of a possession, which includes positions and current trajectories of all the players and the ball, as well as on individual preferences of the players. To predict the trajectories, we propose an uncertainty-aware, multi-modal deep learning model. The model is trained to predict multiple trajectories of a player and probabilities that a given trajectory will be selected. Figure \ref{fig:path_multi}\footnote{ \href{https://drive.google.com/file/d/1gfjI41XmtBZfmKFijayqv60u8CUR0b3S/view}{\color{blue}Link to Supplementary Material} } shows an example of such trajectories and the associated probabilities, compared to baseline models. We provide an in-depth discussion of Figure \ref{fig:paths} in the Results section, and evaluate the proposed method using fine-scale player tracking data collected during several months of an NBA season. In addition, we showcase that with our proposed training regime, the model has the ability of recreating distinct playing styles of individual players.

\section{Related Work}
Modeling and predicting human trajectories is an important challenge in a number of scientific areas. Researchers have worked on this problem to develop realistic crowd simulations \cite{pelechano2007controlling}, or to improve vehicle collision avoidance systems \cite{keller2013will} through predicting future pedestrian movement. When it comes to traffic applications, pedestrian behavior was usually modeled using attracting and repulsive forces to guide them towards a goal, while simultaneously avoiding obstacles. Human pedestrian prediction was also used to improve accuracy of tracking systems \cite{choi2012unified,pellegrini2010improving,yamaguchi2011you} or to study intentions of individuals or groups of people \cite{choi2013understanding,lan2010beyond,xie2017learning}. The advances in deep learning led to data-driven methods, such as Long Short-Term Memory (LSTM) networks \cite{Hochreiter:1997} with shared hidden states \cite{Alahi_2016social}, multi-modal Generative Adversarial Networks (GANs) \cite{Gupta_2018social}, or inverse reinforcement learning \cite{kitani2012activity}, outperforming the traditional methods. The work by \cite{Gupta_2018social} is particularly related to our study, through its use of a multi-modal loss function and by showing practical benefits of multi-modal trajectory prediction as compared to single trajectory predictions. Beyond pedestrian movement, recent research on predictive modeling of vehicular trajectories for self-driving car applications also contains ideas of relevance for the current study. In particular, \cite{Cui_2018_multimodal} showed that multi-modal trajectory predictions for vehicles generate realistic real-world traffic trajectories. The multi-modal loss function in our approach is inspired by this work, where we adapt ideas from the self-driving domain to modeling of movement of basketball players.

The ubiquitous use of tracking systems in professional sports leagues like the NBA or the English Premier League inspired researchers to analyze and model trajectories of athletes during matches. In ECCV 2018, \cite{Felsen_2018_ECCV} used Variational Autoencoders (VAEs) to model real-world basketball data and showed for NBA data that the offensive player trajectories are less predictable than the defense. \cite{Le_2017} and \cite{Seidl_2018} used LSTM to predict near-optimal defensive positions for soccer and basketball, respectively. \cite{sun2019stochastic} similarly used variants of VAEs to generate trajectories for NBA players.  NBA player trajectory predictions are also studied by \cite{Zhan_2019} and \cite{Zheng_2017}, where a deep generative model based on VAE and LSTM and trained with weak supervision was proposed to predict trajectories for an entire team. Macro-intents for each player were inferred, where the players target a spot on the court they want to move to.
The authors evaluate the model mostly by human expert preference studies and show they can outperform the baselines, indicating that RNNs can capture information from observational data in sports. However, their trajectories are usually not smooth and no restrictions are set on the position of a player on consecutive time steps, such that the model may output physically unrealistic trajectories. We consider this state-of-the-art approach in our experiments, and show that it is outperformed by the proposed multi-modal method.

\section{Methodology}

\subsection{Problem Setting}

Recent advancements in optical tracking have made it possible to track the players and the ball during an NBA game with good enough accuracy and temporal resolution to recreate the trajectories of all ten players and the ball during an entire basketball game. This allows us to extract 2-D location ${\boldsymbol \ell}^p_t = [x^p_t, y^p_t]$ of player $p$ at time step $t$, with $p \in \{1, \ldots, 10\}$, as well as 2-D location of the ball at time $t$, ${\boldsymbol \ell}^b_t = [x^b_t, y^b_t]$, where $x$-coordinate represents the length of the field while the $y$-coordinate represents the width, with the origin at the upper left corner (see Figure \ref{fig:paths} for illustration). Using an ordered sequence of previous $L+1$ time steps we can generate  historical trajectory of the $p$-th player as ${\bf h}^p_t = [{\boldsymbol \ell}^p_{t-L}, \ldots, {\boldsymbol \ell}^p_{t}]$, where time steps are equally spaced at an interval of $\Delta_t$. Similarly, we can generate a historical trajectory of the ball as ${\bf h}^b_t = [{\boldsymbol \ell}^b_{t-L}, \ldots,  {\boldsymbol \ell}^b_{t}]$. As a convention, we will assume that the first 5 players represent the team on the offense and the last 5 players the team on the defense. We are interested in predicting future trajectory of $p$-th offensive player, represented as a vector ${\boldsymbol \tau}^p_t = [{\boldsymbol \ell}^p_{t+1}, \ldots,  {\boldsymbol \ell}^p_{t+H}]$, where $H$ is the number of future time steps (or horizon) for which we predict the trajectory. We will assume that the {\it player of interest} (i.e., the offensive player for which we are predicting future trajectory) is denoted by player index $P$.

%\begin{table}[t]
%\parbox{0.45\linewidth}{
%    \centering
%    \caption{Input data for one time step, containing a total of 23 values}
%    \begin{tabular}{lccc}
    %\hline
%    \rowcolor{lightgray}
%    {\bf Input} & {\bf Variable} & {\bf Unit} & {\bf Range} \\ \hline
%    ball & $x$ & [ft] & [0, 94] \\
%     & $y$ & [ft] & [0, 50] \\ \hline
%    \rowcolor{lightgray}
%    player $P$ & $x$ & [ft] & [0, 94] \\
%    \rowcolor{lightgray}
%     & $y$ & [ft] & [0, 50] \\ \hline
%    4 team mates & $x$ & [ft] & [0, 94] \\
%      & $y$ & [ft] & [0, 50] \\ \hline
%     \rowcolor{lightgray}
%     5 opponents & $x$ & [ft] & [0, 94] \\
%     \rowcolor{lightgray}
%      & $y$ & [ft] & [0, 50] \\ \hline
%     shot clock & $t$ & [s] & [0, 24] \\
%     \hline
%     \end{tabular}
%     \label{tab:input}
% }
% \hspace{0.25cm}
% \parbox{0.5\linewidth}{
%     \vspace{-1.9cm}
%     \caption{Output contains a time series of $H$ output velocities in $x$- and $y$-directions, as well as per-mode probability for $M$ modes}
%     \begin{tabular}{lccc}
%     %\hline
%     \rowcolor{lightgray}
%     {\bf Output} & {\bf Variable} & {\bf Unit} & {\bf Range} \\ \hline
%     $M$ trajectories for & $\hat{\boldsymbol \nu}_m$ & [ft/s] & [-25, 25] \\
%     player $P$ & & & \\ \hline
%     \rowcolor{lightgray}
%     $M$ probabilities & $\hat{p}_m$ & [none] & [0, 1] \\ \hline
%     \end{tabular}
%     \label{tab:output}
% }
% \end{table}

In this paper, we processed the raw tracking data to create labeled data set $\mathcal{D} = \{ ({\bf u}_t^P, {\boldsymbol \tau}_t^P), t = 1, \ldots, T, P = 1, \ldots, 5\}$, where one data point is defined for each time step and each offensive player (as indicated by the range $P = 1, \ldots, 5$). Here $T$ is the total number of time steps, input vector ${\bf u}_t^P = \{{\bf h}_t^P, {\bf h}_t^{-P}, {\bf h}_t^b, s_t\}$ is a set of historical player and ball trajectories, where ${\bf h}_t^P$ indicates history of the player of interest, ${\bf h}_t^{-P}$ indicates histories of all other 9 players, and $s_t$ is the shot clock defined as the time in seconds remaining until the shot clock expires. Note that in the input vector the history of the player of interest $P$ always comes first, followed by histories of their $4$ teammates and then by $5$ opposing players, ordered by a distance to the player of interest. Output vector ${\boldsymbol \tau}_t^P$ is a future trajectory of the player of interest $P$ computed at time step $t$, and objective is to build a predictor that accurately predicts their trajectory given inputs ${\bf u}_t^P$.
We emphasize that, in addition to the given inputs, there are other features that potentially might influence the observed trajectories, such as game clock, home vs. away, foul calling, previous plays, or player mismatch. As we demonstrate with the shot clock feature, our approach allows for a straightforward use of any additional feature that a modeller may deem important. However, an in-depth feature analysis is out of scope of this paper, and instead we focus on showing viability of the proposed multi-modal predictive model. In fact, it could be argued that a number of such features are implicitly present in the input representation already. For example, if a team has a large point lead with little game time remaining, they may slow down on the offense and the observed movement history could capture that information.

Lastly, note that an alternative to predicting a sequence of $H$ future locations of the offensive player is predicting a sequence of their velocities. As we know the current location at time $t$, we can convert trajectory ${\boldsymbol \tau}_t^P$ to a velocity vector ${\boldsymbol \nu}_t^P = [{\bf v}_{t+1}^P, \ldots, {\bf v}_{t+H}^P]$ using a direct mapping of velocities to locations, computed for horizon $h \in \{1, \ldots, H\}$ as
\begin{equation}
\begin{split}
\label{eq:conversion_t2v}
{\bf v}_{t+h}^P = & [v^P_{x,t+h}, v^P_{y,t+h}] = \\
 & [\frac{x^P_{t+h} - x^P_{t+h-1}}{\Delta_t}, \frac{y^P_{t+h} - y^P_{t+h-1}}{\Delta_t}].
\end{split}
\end{equation}
Although trajectories and velocity vectors are mathematically interchangeable, a particular choice might have a significant impact on model training. As we will demonstrate experimentally, predicting the next location is more challenging due to the issue in normalization of coordinates. 

\subsection{Proposed Approach}
As noted previously \cite{Zhan_2019}, movement of basketball players is inherently multi-modal as the players can decide between multiple plausible trajectories at any given time (e.g., to move towards the basket for a layup or towards a corner for a three-point attempt). In order to account for this multi-modality we train a predictive model that generates output ${\bf o}^P_t = [\hat{{\boldsymbol \nu}}_{t,1}^P, \ldots, \hat{{\boldsymbol \nu}}_{t,M}^P, \hat{p}_{t,1}^P, \ldots, \hat{p}_{t,M}^P]$, which consists of $M$ predicted trajectories $\hat{{\boldsymbol \nu}}_{t,m}^P$ representing $M$ modes, as well as $M$ scalars $\hat{p}_{t,m}^P$ representing probabilities that a corresponding mode is selected by a player. This results in $(2 H + 1) M$ output values, since output for each mode consists of a trajectory comprising $H$ 2-D locations and an additional mode probability. 
%The model outputs are summarized in Table \ref{tab:output}. 

\subsubsection{Loss function}
Given a ground-truth trajectory ${\boldsymbol \nu}$ and predicted trajectory $\hat{{\boldsymbol \nu}}$, we first define the trajectory loss as
\begin{equation}
\label{eq:MSE_loss}
\mathcal{L}^{\mathrm{MSE}}({\boldsymbol \nu}, \hat{\boldsymbol \nu}) = \frac{1}{2 H} \|{\boldsymbol \nu}-\hat{\boldsymbol \nu}\|_2^2,
\end{equation}
defined as a mean squared error (MSE) of the predicted velocity vector.  
Then, in order to train a model to predict multiple trajectories and their probabilities, we base our approach on an adaptation of the multi-modal loss function presented in \cite{Cui_2018_multimodal}. A similar loss function is used by \cite{Gupta_2018social} to generate multi-modal pedestrian trajectories within a GAN framework. 
%Intuitively, for each training example we want to select the best matching mode and update it through backpropagation, while leaving the non-matching modes unchanged. Following this intuition, 
In particular, we define the Multiple-Trajectory Prediction (MTP) loss for time step $t$ and player $P$, comprising a linear combination of classification loss $\log \hat{p}_m$ and trajectory loss \eqref{eq:MSE_loss},
\begin{equation}
\label{eq:MTP_loss}
\mathcal{L}^{\mathrm{MTP}} = \sum_{m=1}^M \delta_\epsilon(m = m^*) \Big( \log \hat{p}_m + \alpha  \mathcal{L}^{\mathrm{MSE}}({\boldsymbol \nu}_{t}^P, \hat{\boldsymbol \nu}_{t,m}^P) \Big),
\end{equation}
where $\hat{p}_m$ is an output of a softmax, $\alpha$ is a hyper-parameter used to trade-off the classification and trajectory losses, and $m^*$ is the index of the winning mode that produced the trajectory closest to the ground truth, computed according to a distance function $dist()$ defined in the next subsection,
\begin{equation}
\label{eq:best_m}
m^* = \argmin_{m \in \{1, \ldots, M\}} dist({\boldsymbol \nu}_{t}^P, \hat{\boldsymbol \nu}_{t,m}^P).
\end{equation} 
Moreover, $\delta_\epsilon$ is a relaxed Kronecker delta \cite{Rupprecht_2017} giving the most weight to the best matching trajectory, but also a small weight to the remaining ones,
\begin{equation}
\label{eq:binary_indicator}
\delta_\epsilon(cond) = \begin{cases}
    1-\epsilon, & \text{if condition $cond$ is true},\\
    \frac{\epsilon}{M-1}, & \text{otherwise}.
  \end{cases}
\end{equation}
Intuitively, the classification loss in \eqref{eq:MTP_loss} forces the probability of the winning mode to 1 (thus pushing probabilities of other modes towards zero due to the softmax), and trajectory loss penalizes prediction error of the winning mode.

We note that \cite{Cui_2018_multimodal} used the unrelaxed Kronecker delta (i.e., $\epsilon$ was set to $0$), which only updates the closest trajectory. In practice, this leads to problems where a randomly initialized path is much worse than the remaining paths. Such poorly initialized modes never get selected through \eqref{eq:best_m} and do not get a chance to improve during training. To prevent this issue we use the relaxed Kronecker delta, where we start from some small value of $\epsilon$ that is gradually reduced towards $0$ as the training progresses. This phenomenon is well known in generative models and is commonly called mode collapse in GANs or posterior collapse in VAEs. Comparable annealing remedies have been proposed in VAEs \cite{bowman2015generating}, but are generally not sufficient to achieve good performance \cite{fu2019cyclical}. Our approach was more stable than VAE or GAN training and we will empirically show that we can outperform state-of-the-art models based on each of those two methods.

\subsubsection{Distance functions}
As mentioned above, $m^*$ denotes a path closest to the ground truth, however there are different closeness measures that can be considered. For example, in \cite{Gupta_2018social} the closest mode is defined simply as a path with the lowest trajectory loss, computed as
\begin{equation}
\label{eq:dist_mse}
dist_{MSE}({\boldsymbol \nu}, \hat{\boldsymbol \nu}_m) = \mathcal{L}^{\mathrm{MSE}}({\boldsymbol \nu}, \hat{\boldsymbol \nu}_m) .
\end{equation}
We also considered other distance functions, as \cite{Cui_2018_multimodal} concluded that its choice has a large impact on the model performance. Thus, we considered distance function with the smallest overall displacement error, defined as a location error at the last time step and computed as
\begin{equation}
\label{eq:dist_l}
dist_l({\boldsymbol \nu}, \hat{\boldsymbol \nu}_m) = \| \sum_{h=1}^{H} ({\boldsymbol \nu}_{t+h} - \hat{\boldsymbol \nu}_{t+h, m}) \|_2 .
\end{equation}
Lastly, we considered using the error of final player velocity (which can be interpreted as player's ``heading"), shown in earlier work \cite{Cui_2018_multimodal} to be beneficial,
\begin{equation}
\label{eq:dist_v}
dist_v({\boldsymbol \nu}_t, \hat{\boldsymbol \nu}_{t, m}) = \norm{ {\boldsymbol \nu}_{t+H} - \hat{\boldsymbol \nu}_{t+H, m} }_2.
\end{equation}

\subsubsection{Neural Network Architecture}

While \cite{Cui_2018_multimodal} use the multi-modal loss function to train a CNN model, we will show that on the NBA data LSTM network is more effective. We use a two-layer LSTM architecture, each with a width of 128, to encode the time-series input of recently observed data ${\bf u}_t^P$. The encoder is a fully connected layer and the prediction consists of $M$ trajectories of a single player given as $x$- and $y$-velocities for $H$ future time steps, as well as $M$ probabilities that the player will follow the respective trajectory. 
It is important to note that the players differ in their positions, skills, heights, and weights, and we would expect them to run at different speeds and along different paths. To take these differences into account we consider a two-stage training approach to learn specific per-player models. To this end we first train the proposed model on data taken from all players, which can be seen as learning average behavior of all NBA players. Then, in the second training phase these pre-trained networks can be used to initialize a specialized per-player network fine-tuned on a subset containing only that player's data, so that individual behavior of the player can be learned. In the experiments we evaluate both global and per-player models.

We refer to the proposed multi-modal approach as Multi-modal Basketball Trajectories (MBT). We evaluate different number of modes $M$ and investigate different distance functions in \eqref{eq:best_m}, indicating these choices in the subscript. In particular, we denote model variants as MBT$_{Md}$, with $d \in \{MSE, l, v\}$, corresponding to \eqref{eq:dist_mse}, \eqref{eq:dist_l}, and \eqref{eq:dist_v}, respectively. For example, MBT$_{4l}$ generates 4 paths and uses distance function \eqref{eq:dist_l} during training. When using a single mode the distance measure is not used, and we refer to the uni-modal model as MBT$_1$. %(although this approach does not produce multiple paths).

\section{Experiments}

\subsubsection{Data set}

We used publicly available movement data collected from 632 NBA games during the 2015-2016 season\footnote{\href{https://github.com/sealneaward/nba-movement-data}{\color{blue}https://github.com/sealneaward/nba-movement-data}, last accessed June 2020; we are not associated with the data creator in any way.}, from which we extracted 114,294 offensive possessions. An offensive possession starts when all players from one team cross into the opponents' half court, and ends when the first offensive player leaves the half court or the game clock is paused. Possessions shorter than $3s$ were discarded, resulting in 113,760 possessions. This amounts to 1.1 million seconds of gameplay where player location is captured every $0.04s$. We downsampled the data by a factor of 3 to obtain sampling rate of $\Delta_t=0.12s$, corresponding to a lower bound on human reaction time \cite{Fischer:1984} during which velocity is considered constant.
Furthermore, we randomly split the data into train and test sets using 90/10 split. All inputs and outputs were normalized to the $[-1, 1]$ range.
To train the specialized networks that predict specific player's movement we extracted possessions featuring that player. The amount of data for each player is in the order of several thousands (e.g., for Stephen Curry there were 2,767 possessions).

\subsubsection{Model training}
As discussed previously, we used a 2-layer LSTM with 128 channels in each layer. To learn the general model for all NBA players we trained LSTM in batches of 1,024 samples. The learning rate in Adam optimizer was set to $5 \cdot 10^{-4}$. We set hyper-parameter $\alpha$ in equation \eqref{eq:MTP_loss} to $1$, such that the amplitude of the two losses are about equal, and $\epsilon$ in \eqref{eq:binary_indicator} to $0.25$ which was reduced by a factor of $0.05$ per epoch until $\epsilon=0$. We used $\ell_2$ regularization with the weight of $\lambda = 0.001$ and an early stopping mechanism to further prevent overfitting.
To specialize the neural network for a specific player we fine-tune the base model on data from that player. We start with $\epsilon = 0.75$ which is reduced by a factor of 0.01 per epoch to make sure that all modes benefit from the information contained in this smaller training set. The initial learning rate in this case was reduced to $10^{-5}$.

All training was done on a single computer with Nvidia GeForce GTX 1080 card. It took approximately 60 minutes to train the base model, while specializing the network on a specific player took less than 5 minutes.

\subsubsection{Accuracy measures}

We report common measures used in pedestrian trajectory prediction, \textit{final displacement error} (FDE) and \textit{average displacement error} (ADE) \cite{Alahi_2016social,Gupta_2018social},
defined as 
\begin{equation}
\begin{split}
\label{eq:FDE/ADE}
\text{FDE} = & \frac{1}{5T} \sum_{t=1}^T \sum_{P=1}^5 \norm{{\boldsymbol \ell}^P_{t+H} - \hat{\boldsymbol \ell}^P_{t+H}}_2, \\
\text{ADE} = & \frac{1}{5HT} \sum_{t=1}^T \sum_{P=1}^5 \sum_{h=1}^H \norm{{\boldsymbol \ell}^P_{t+h} - \hat{\boldsymbol \ell}^P_{t+h}}_2.
\end{split}
\end{equation}
In other words, FDE considers the location error at the end of the prediction horizon $H$, while ADE averages location errors over the entire trajectory.
We also report MSE error, defined as in equation \eqref{eq:MSE_loss}. Unlike FDE and ADE that measure trajectory prediction errors, MSE is a measure of how accurately are the velocities predicted.

To evaluate multi-modal approaches we choose the path that has the smallest FDE among all the generated paths, which is consistent with evaluation procedure used in the literature \cite{Gupta_2018social,Rupprecht_2017}.

\subsubsection{Baselines}
To establish an upper bound for the proposed error measures we compared our method to two straw-man baselines. \textit{Constant location} (CL) baseline assumes that a player stays at the last observed location on the court, while \textit{constant velocity} (CV) baseline assumes that the player keeps moving in the last observed direction with constant speed.

Baseline \textit{CNN} refers to an approach that transforms the input to a rasterized trace image and uses a CNN encoder (instead of LSTM) before predicting the future velocities \cite{Cui_2018_multimodal}. For the encoder, we used 5 layers with depths [64, 128, 128, 64, 32], 5x5 mask, "same" padding, and 2x2 max pooling. The decoder consisted of 2 densely connected layers with size 128 and 64.

\begin{table*}[!ht]
\centering
\caption{Comparison of various models, input steps $L$, and modes M in terms of error metrics ADE and FDE (in feet) and MSE (in ft$^2$/s$^2$)}
{ \small
  \begin{tabular}{crr  rrr  rrr  rrr}
    &  & {}  & \multicolumn{3}{c}{\bf ${\boldsymbol H}$ = 10} & \multicolumn{3}{c}{\bf ${\boldsymbol H}$ = 20}  & \multicolumn{3}{c}{\bf ${\boldsymbol H}$ = 40}  \\
    {\bf Method} & {${\boldsymbol L}$} & ${\boldsymbol M}$ & {\bf ADE}  & {\bf FDE} & {\bf MSE} & {\bf ADE} & {\bf FDE} &  {\bf MSE} & {\bf ADE}  & {\bf FDE} &  {\bf MSE} \\
    \hline
    \rowcolor{lightgray} 
    CL & 1 & 1 & 3.19 & 5.69 & 39.56 & 5.47 & 9.78 & 38.46 & 9.03 & 14.85 & 37.34 \\
    \rowcolor{lightgray} 
    CV & 1 & 1 & 1.72 & 3.92 & 9.09 & 4.64 & 10.97 & 16.01 & 11.59 & 26.14 & 20.59 \\
    % AL & & 16.529 & 16.516 & & 16.682 & 16.603 & & 16.752 & 16.650 & \\
    \rowcolor{lightgray} 
    CNN & 10 & 1 & 2.76 & 5.25 & 15.80 & 5.28 & 9.99 & 17.48 & 8.15 & 13.23 & 21.95 \\
    \rowcolor{lightgray} 
    location-LSTM & 10 & 1 & 1.61 & 2.98 & 10.21 & 3.43 & 6.91 & 15.94 & 6.79 & 12.11 & 29.80 \\
    \hline
    MBT$_1$ & 10 & 1 & 1.43 & 2.98 & 7.26 & 3.32 & 6.92 & 12.36 & 6.59 & 11.97 & 16.93 \\
    MBT$_1$ & 20 & 1 & 1.40 & 2.93 & 7.25 & 3.30 & 6.91 & 12.41 & 6.59 & 11.97 & 16.74 \\
    MBT$_1$ & 30 & 1 & 1.39 & 2.92 & 7.46 & 3.33 & 6.91 & 12.32 & 6.58 & 11.92 & 16.87 \\
    \hline
    \rowcolor{lightgray}
    SocialGAN$_1$ & 10 & 1 & 1.25 & 2.75 & 8.18 & 3.09 & 6.67 & 13.32 & 6.47 & 12.35 & 17.54 \\
    \rowcolor{lightgray}
    MACRO VRNN$_1$ & 10 & 1 & 1.70 & 3.43 & 13.17 & 4.46 & 8.66 & 19.85 & 8.48 & 14.98 & 25.03 \\
    \rowcolor{lightgray}
    SocialGAN$_4$ & 10 & 4 & 1.19 & 2.61 & 7.36 & 2.95 & 6.33 & 11.91 & 6.19 & 11.54 & 15.76 \\
    \rowcolor{lightgray}
    MACRO VRNN$_4$ & 10 & 4 & 1.07 & 1.98 & 5.90 &3.14 & 5.07 & 11.93 & 6.40 & 8.54 & 19.29 \\
    \hline
    MBT$_{4MSE}$ & 10 & 4 & \bf{1.01} & 1.91 & \bf{3.82} & \bf{2.33} & \bf{4.00} & \bf{6.35} & 5.25 & 6.92 & 12.46 \\
    MBT$_{4v}$ & 10 & 4 & 1.05 & 1.93 & 4.00 & 2.66 & 4.31 & 7.75 & 6.71 & 8.74 & 14.72 \\
    MBT$_{4l}$ & 10 & 4 & \bf{1.01} & \bf{1.90} & \bf{3.82} & \bf{2.33} & 4.04 & \bf{6.35} & \bf{4.89} & \bf{6.39} & \bf{11.56} \\
    \hline
\end{tabular}
}
\label{tab:results}
\end{table*}

To compare different output alternatives we trained the same LSTM architecture used for our model to directly predict player locations, as opposed to predicting velocities. We refer to this model as the \textit{location-LSTM}.
We also considered \textit{SocialGAN} \cite{Gupta_2018social}, the state-of-the-art in human trajectory prediction. This approach uses an LSTM-based generator, coupled with a social pooling layer to account for nearby actors. We trained this model using code made available by the original authors\footnote{\href{https://github.com/agrimgupta92/sgan}{\color{blue}https://github.com/agrimgupta92/sgan}, last accessed June 2020.}, using the same NBA data set except that SocialGAN can not use extra information such as ball location or shot clock, therefore only the players trajectories are used. GANs are notoriously hard to train, which resulted in a training time of 28 hours for 50 epochs of training.
In addition, we considered the state-of-the-art MACRO VRNN \cite{Zhan_2019}, which uses programmatic weak supervision to first predict a location that the player wants to reach and then uses a Variational RNN (VRNN) to predict a trajectory that the player will take to reach it. MACRO VRNN also accounts for the multi-modality of the problem, with the number of generated paths denoted in the subscript.
We used models provided in \cite{Zhan_2019} trained on roughly the same amount of data. Note that training takes up to 20 hours, as opposed to only 1 hour for our proposed method. 

\subsection{Results}

We first compared performance of models trained on data containing all possessions, with results across different error measures and time horizons presented in Table \ref{tab:results}. 

Model CL predicts that the player will remain static at the last observed location, which explains the large MSE. CL also has a relatively large FDE for shorter horizons, but does not deteriorate as fast as the CV model which assumes the player will keep moving with the last observed velocity. The CNN model outperformed the simple baselines at longer horizons, however at short horizons the performance was suboptimal. Location-LSTM (which predicts player's locations instead of velocities as the competing methods) is comparable to MBT$_1$ model in terms of ADE and FDE metrics, with much worse MSE metric. As we will demonstrate later in qualitative results, this difference in MSE can be explained by the fact that location-LSTM produces trajectories that are not physically achievable by the players.

Next we experimented with the uni-modal MBT$_1$ model and evaluated the influence of different lengths of historical inputs $L$. Based on the results we confirmed that the MBT$_1$ models only marginally improve with longer input sequences. As a result, in the remainder of the experiments we use a value of $L=10$, consistent with \cite{Zhan_2019}.

In the following experiment we compared different distance functions used for training MBT methods, where we kept $M$ fixed at $4$. We see that the choice of distance function had limited effect on accuracy measures at a shorter horizon of $1.2s$. However, as the horizon increased, MBT$_{4l}$ started outperforming the competing approaches by a considerable margin. Taking this result into account, in further experiments we used the distance function defined in \eqref{eq:dist_v}.

When we compare the proposed method to the state-of-the-art models MACRO VRNN and SocialGAN, we separate the analysis by comparing the same number of modes. When evaluating a single trajectory, SocialGAN outperforms both our approach and MACRO VRNN in ADE and FDE. However, MBT$_1$ reaches better MSE than those approaches. When comparing multiple modes, we can see that MBT$_{4l}$, MBT$_{4v}$ and MBT$_{4MSE}$ performance is roughly comparable at shorter horizons, but MBT$_{4l}$ outperforms all other methods across all accuracy measures at longer horizons. Quite notably, MBT$_{4l}$ outperforms the baselines with a large margin in terms of MSE velocity measure. For example, for horizon $H = 40$, our MBT$_{4l}$ model achieves ADE $24\%$ and $21\%$ smaller than MACRO VRNN$_4$ and SocialGAN$_4$, respectively.

\begin{figure}[ht]
\begin{center}
\includegraphics[width=0.9\linewidth]{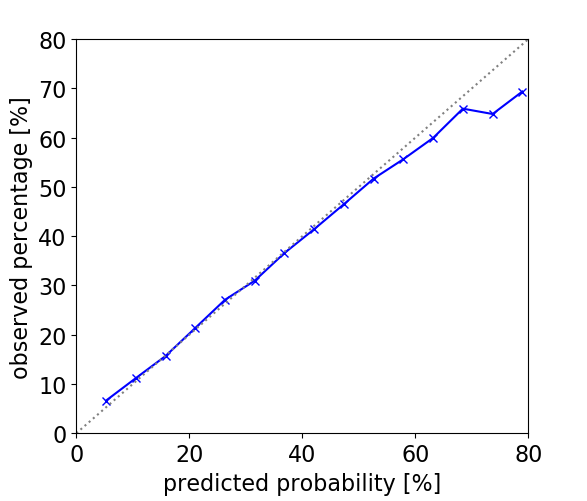}
\end{center}
%\vspace{-2.5mm}
\caption{Evaluation of predicted mode probabilities for MBT$_{4l}$}
\label{fig:calibration}
\end{figure}

%We also evaluated quality of inferred mode probabilities produced by the MBT$_{4l}$ model. To this end we compared predicted mode probabilities to empirical ones, computed as a frequency of how often a mode of certain probability had the lowest FDE. We bucketed inferred probabilities in 5\% bins and for each computed the empirical probability,  with the average per-bucket results presented in Figure \ref{fig:calibration}. We can see that the plot closely follows the identity line, indicating that the predicted mode probabilities are well-calibrated.

In Figure \ref{fig:paths}, we illustrate predicted trajectories for a randomly picked player. Trajectories were generated using a single-path model that predicts locations (location-LSTM, Figure \ref{fig:path_loc}), two single-path models that predicts velocities, one based on a CNN architecture (Figure \ref{fig:path_CNN}) and one based on an LSTM architecture (MBT$_1$, Figure \ref{fig:path_vel}), one sample path of MACRO VRNN (Figure \ref{fig:path_VRNN}), 4 sampled paths of SocialGAN (Figure \ref{fig:path_SGAN}) and our proposed method for 4 modes MBT$_{4l}$ (Figure \ref{fig:path_multi}). We can see that location-LSTM output is noisy and does not represent realistic player movements. Player trajectories predicted by the CNN and MBT$_1$ model are smoother and more realistic, showing the advantage of predicting velocities instead of locations. While CNN and MBT$_1$ generate qualitatively similar results, MBT$_1$ outperforms CNN in the quantitative measures. MACRO VRNN generally produces paths that are less smooth than competing models, explaining the high error in MSE as discussed above. The multiple paths predicted by SocialGAN are smooth and look plausible, but lack the diversity of movement that we would basketball trajectories. MBT$_{4l}$ predicts 4 paths that are very distinct from each other. The highest-probability path ends up very close to the observed final player location, while accurately following the ground-truth trajectory. Other paths produced by the multi-modal model allow for diverse movements, such as an aggressive drive to the basket or supporting the ball-handling teammate near the center of the court.

We also evaluated quality of inferred mode probabilities produced by the MBT$_{4l}$ model. To this end we compared predicted mode probabilities to empirical ones, computed as a frequency of how often a mode of certain probability had the lowest FDE. We bucketed inferred probabilities in 5\% bins and for each computed the empirical probability,  with the average per-bucket results presented in Figure \ref{fig:calibration}. We can see that the plot closely follows the identity line, indicating that the predicted mode probabilities are well-calibrated.

To evaluate the hypothesis that the MBT trajectories are more physically realistic, we calculated acceleration of predicted trajectories on the test set. The maximum acceleration of MBT$_{4l}$ is $12.2 m/s^2$. We note that the ground truth contains noisy outliers, with accelerations of up to $600 m/s^2$ (the $99.9$th percentile is $14.5 m/s^2$). In contrast, when considering MACRO VRNN we observe accelerations of more than $500 m/s^2$ (the $99.9$th percentile is $54.86 m/s^2$). This indicates that in many cases the baseline trajectories are far from being physically achievable, while the proposed method yielded more realistic outputs.

\subsubsection{Evaluation of per-player models}

\begin{table} [t]
\small
\centering
\caption{Prediction of specific players with and without fine-tuning for $H = 40$ (4.8 seconds) using the MBT$_{4l}$ model}
\vspace{-2mm}
{
  \begin{tabular}{ccc c cc c cc c c}
    {\bf Player} & {\bf Fine-tuned?}  & {\bf ADE}  & {\bf FDE} & {\bf MSE}  \\
    \hline
    \rowcolor{lightgray} 
    LeBron James & No & 4.78 & 6.63 & 9.97 \\
    \rowcolor{lightgray} 
    LeBron James& Yes & 4.67 & 6.24 & 9.91 \\
    \hline
    Stephen Curry & No & 6.32 & 7.80 & 17.35 \\
    Stephen Curry & Yes & 6.09 & 7.51 & 16.62 \\
    \hline
    \rowcolor{lightgray} 
    Russell Westbrook & No & 5.49 & 7.15 & 12.43 \\
    \rowcolor{lightgray} 
    Russell Westbrook & Yes & 5.36 & 6.90 & 12.23 \\
    \hline
    DeAndre Jordan & No & 4.36 & 6.01 & 12.20 \\
    DeAndre Jordan & Yes & 3.93 & 4.94 & 12.56 \\
    \hline
    \rowcolor{lightgray} 
    Andrew Bogut & No & 4.54 & 6.12 & 9.34 \\
    \rowcolor{lightgray} 
    Andrew Bogut & Yes & 4.29 & 5.40 & 9.03 \\
    \hline
\end{tabular}
}
\label{tab:players}
\vspace{-5mm}
\end{table}

In this section we compare per-player models to the base model trained on all players, as well as the per-player models fine-tuned on players that are playing in the same position, but are known to have distinct playing styles. We first compared performance of the base and per-player models for several example players, with results presented in Table \ref{tab:players}. We can see that per-player models resulted in improved performance across the board, as they are better capturing playing styles of individual players.

\begin{figure*}[t]
\begin{subfigure}{0.32\textwidth}
  \includegraphics[width=\linewidth]{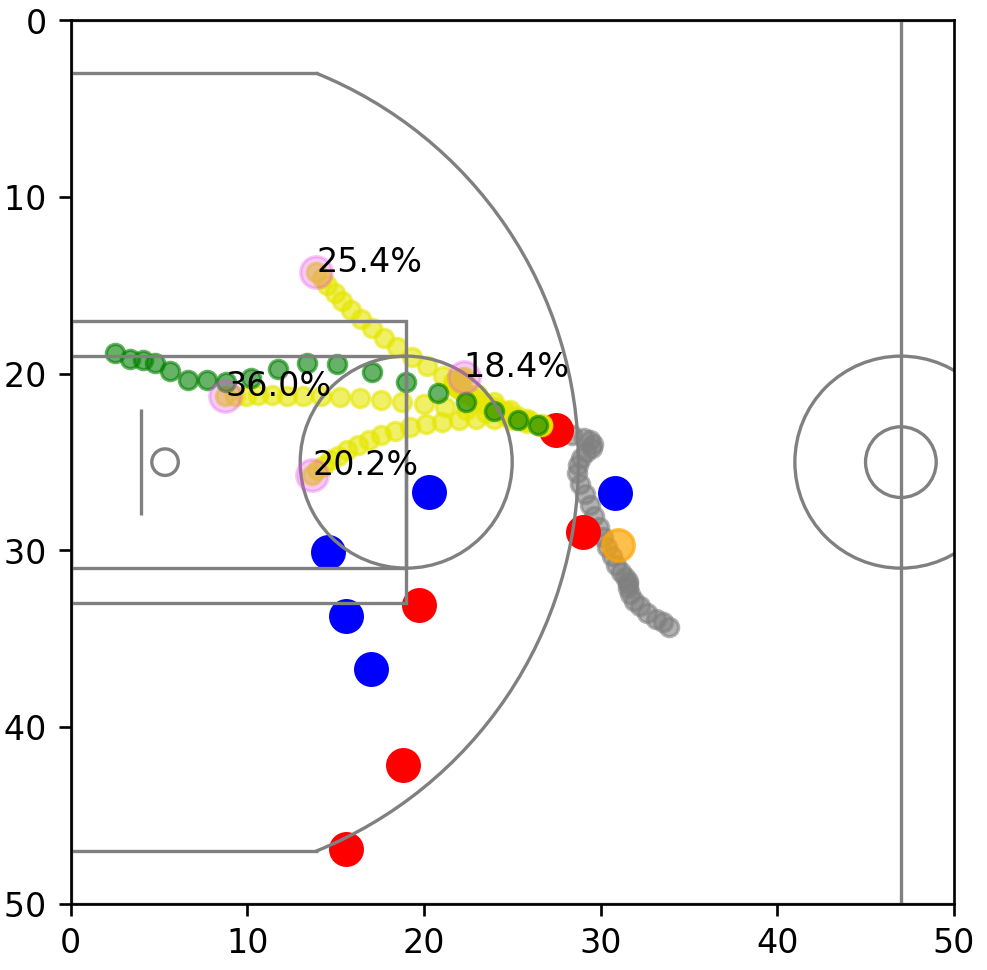}
  \vspace{-5mm}
  \caption{}\label{fig:DeAndre_a}
\end{subfigure}\hfill
\begin{subfigure}{0.32\textwidth}
  \includegraphics[width=\linewidth]{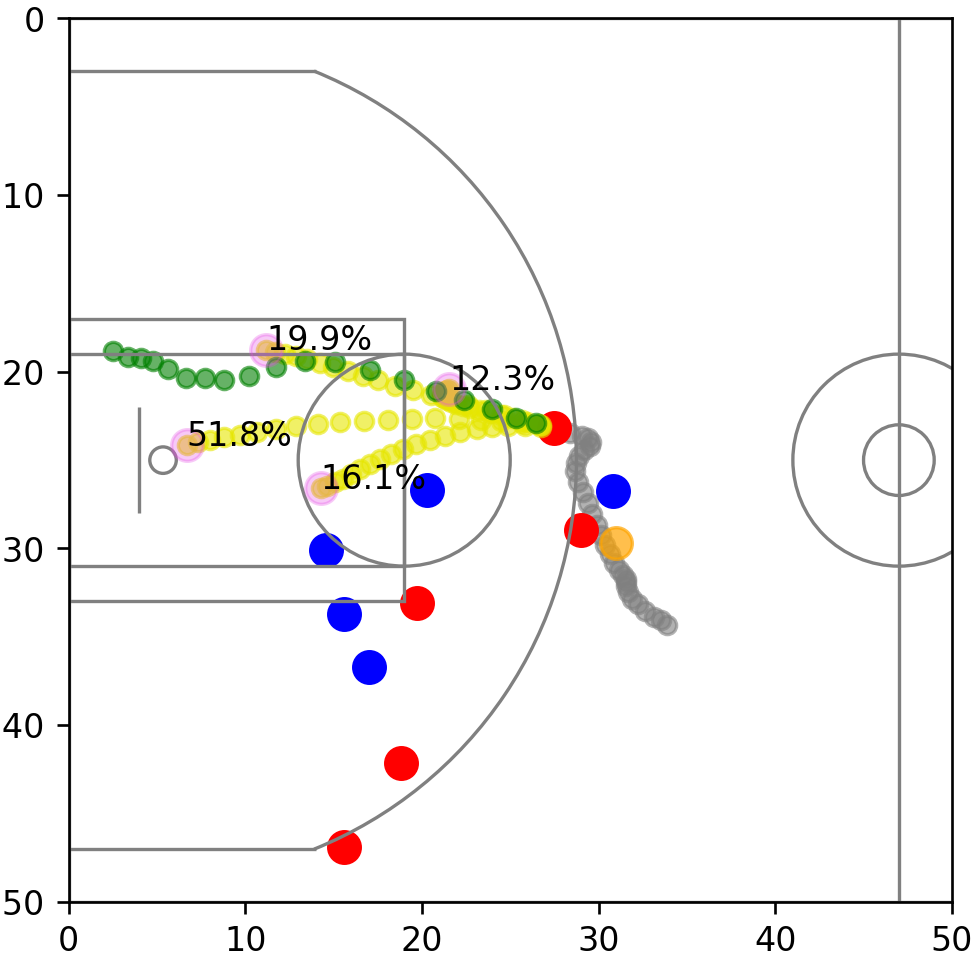}
  \vspace{-5mm}
  \caption{}\label{fig:DeAndre_b}
\end{subfigure}\hfill
\begin{subfigure}{0.32\textwidth}%
  \includegraphics[width=\linewidth]{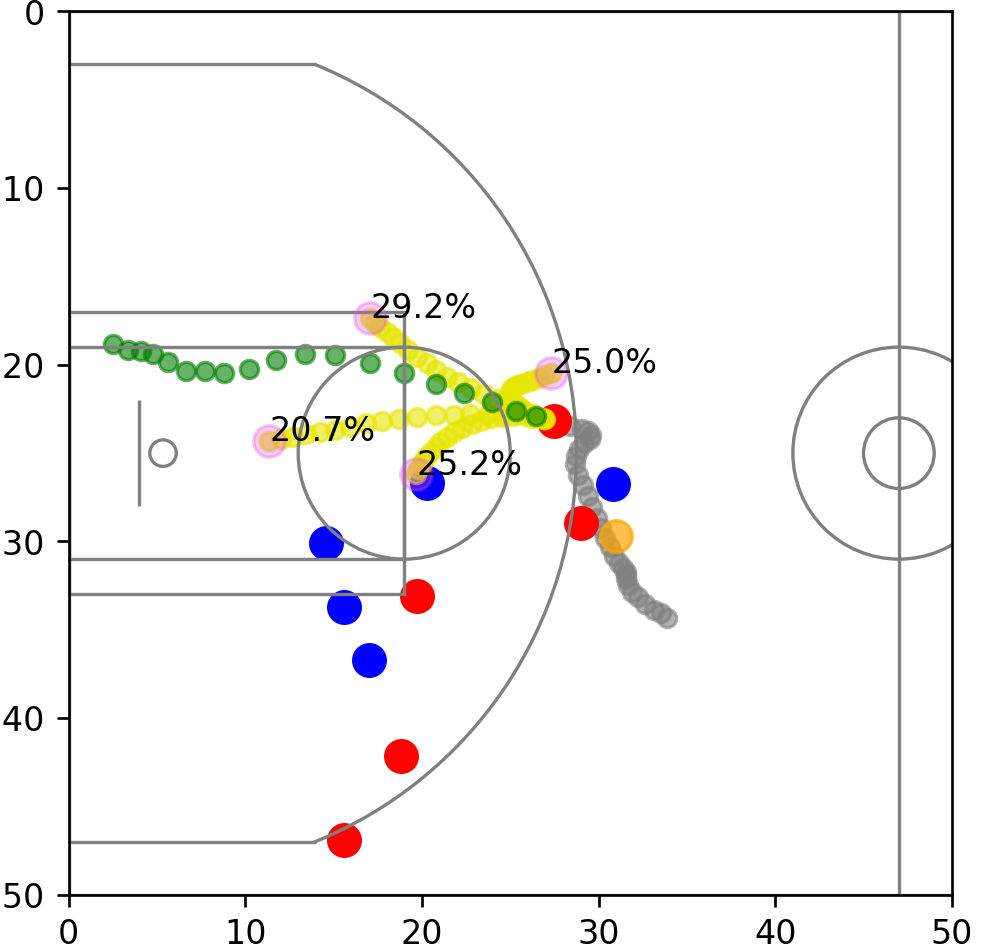}
  \vspace{-5mm}
  \caption{}\label{fig:DeAndre_c}
\end{subfigure}
  \vspace{-2.5mm}
\caption{Visualization of predicted trajectories for DeAndre Jordan with $H=20$ (2.4s) using 3 different networks MBT$_{4l}$: a) trained on all players, b) retrained with the data of DeAndre Jordan and c) retrained with the data of Andrew Bogut %; red: attackers, blue: defenders, orange: ball, grey: input history of predicted player, yellow: prediction, pink: end position, green: ground truth
}
% \vspace{-2.5mm}
\label{fig:DeAndre}
\end{figure*}

\begin{figure*}[t]
\begin{subfigure}{0.32\textwidth}
  \includegraphics[width=\linewidth]{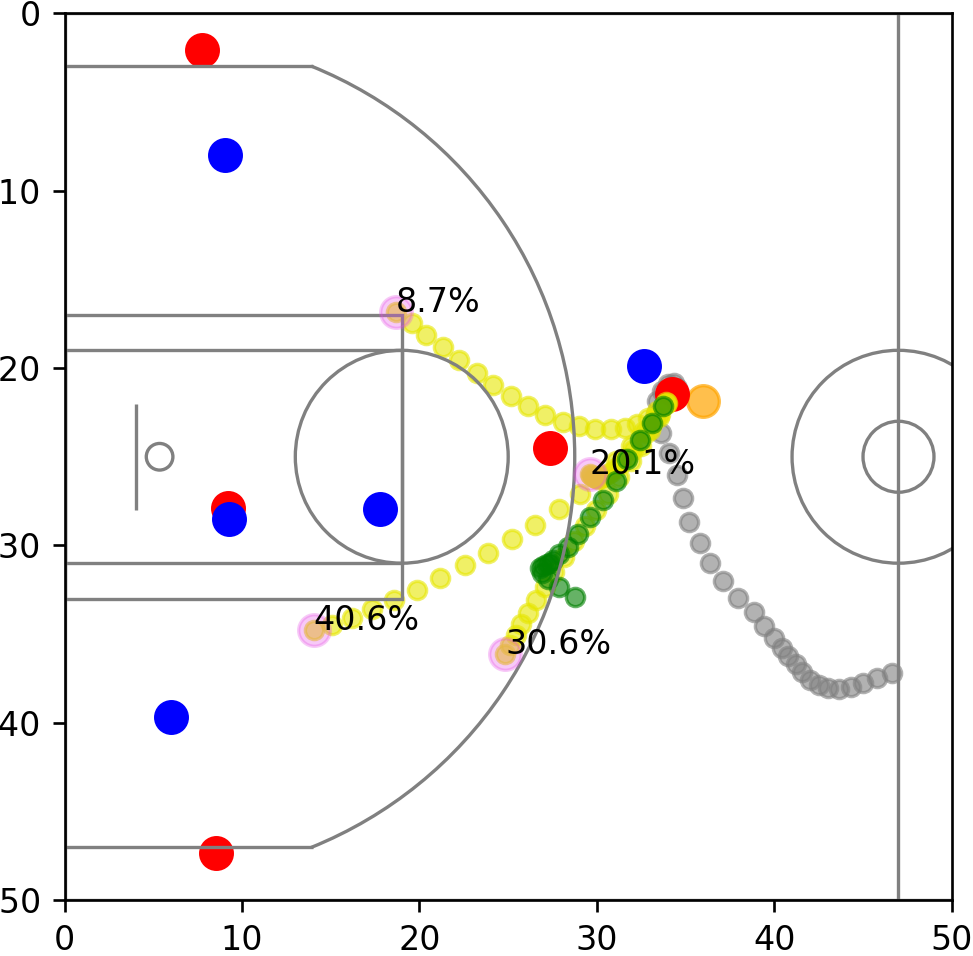}
  \vspace{-5mm}
  \caption{}\label{fig:Curry_a}
\end{subfigure}\hfill
\begin{subfigure}{0.32\textwidth}
  \includegraphics[width=\linewidth]{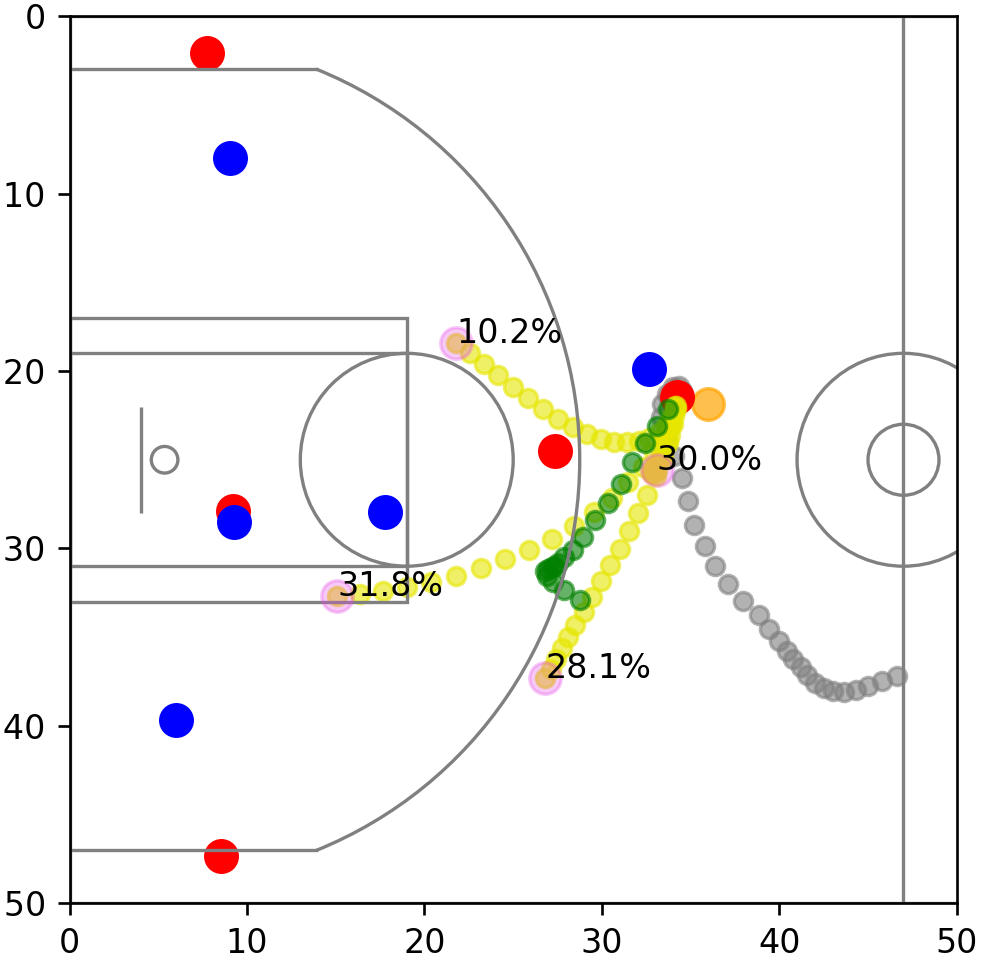}
  \vspace{-5mm}
  \caption{}\label{fig:Curry_b}
\end{subfigure}\hfill
\begin{subfigure}{0.32\textwidth}%
  \includegraphics[width=\linewidth]{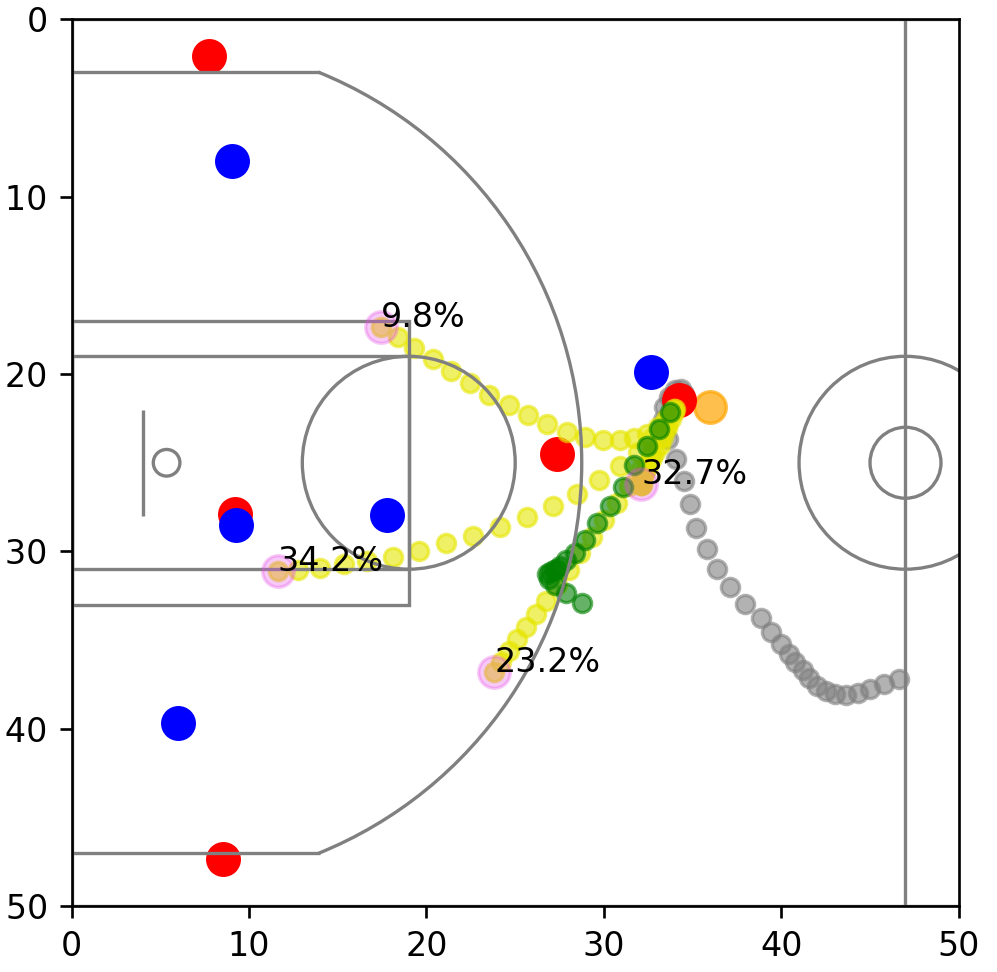}
  \vspace{-5mm}
  \caption{}\label{fig:Curry_c}
\end{subfigure}
\vspace{-2.5mm}
\caption{Visualization of predicted trajectories for Stephen Curry with $H=20$ (2.4s) using 3 different networks MBT$_{4l}$: a) trained on all players; b) retrained with the data of Stephen Curry; c) retrained with the data of Russel Westbrook}
\label{fig:Curry}
%\vspace{-2.5mm}
\end{figure*}

Let us consider a specific game situation where center DeAndre Jordan just set up a pick and roll, shown in Figure \ref{fig:DeAndre} and the animated video in Supplementary Material\footnote{ \href{https://drive.google.com/file/d/1gfjI41XmtBZfmKFijayqv60u8CUR0b3S/view}{\color{blue}Link to Supplementary Material} }. The model trained on all players predicts that the so-called roll man will now move either towards the basket or towards the wide open space on the right-hand side of the court, shown in the first row of Figure \ref{fig:DeAndre_a}. Jordan is a very dynamic and fast center who executes many successful pick and rolls, so our model trained on his data predicts he will drive to the basket faster and with a higher probability than an average player in the same situation, as shown in Figure \ref{fig:DeAndre_b}. We also compared to a model trained on data of Andrew Bogut, a defense specialist who is not as fast as Jordan. According to {\it stats.nba.com}\footnote{\href{https://on.nba.com/2ulXVau}{\color{blue}https://on.nba.com/2ulXVau}, last accessed June 2020.}, Bogut only attempts 0.5 pick and rolls per game, while Jordan attempts 2.4. Our model correctly predicts Bogut's paths to be less dynamic and gives a 25\% probability that he would turn around and focus on defending a counter attack, entirely relying on his team mate to capitalize on the pick, shown in Figure \ref{fig:DeAndre_c}.

The following experiment involves a situation where Stephen Curry has possession of the ball at the top of the circle, with a defender to his right, as illustrated in Figure \ref{fig:Curry} (and in the Supplementary Material). This example shows some limitations of our approach because in actuality Curry first acts like he wants to drive inside, but decides to stop and throw the ball for a 2-pointer before starting to move backwards. The predicted trajectories are much simpler, but still capture some interesting options that the player may choose. The model that was trained on all players predicts that the player may move towards the basket with about 40\% probability as seen in Figure \ref{fig:Curry_a}, with other lower-probability options to move along the arc, stay at the top of the arc, or try to circle around the defender. The model that was retrained on data of Stephen Curry shown in Figure \ref{fig:Curry_b} slightly adjusts the path along the arc, because Curry often tries to shoot 3-pointers (more specifically, he has the second-most 3-point attempts in the 2015/16 season). As a result the model also gives him a lower probability to drive towards the basket. We evaluated the same situation with a network fine-tuned on data of Russell Westbrook, shown in Figure \ref{fig:Curry_c}. Westbrook attempts much fewer 3-pointers than Curry, and instead has more 2-point attempts. He is also a very dynamic player that is excellent at driving to the basket, such that when he makes an attempt he usually gets closer to the basket than an average player would. Thus, when he moves along the arc our model predicts that he will not stay behind the 3-point line, but will instead try to get closer to the basket. We can see the model successfully managed to capture characteristics of individual players, adjusting the predictions to their own playing styles.

\vspace{-2.5mm}
\section{Conclusion}

In this paper we proposed an LSTM-based model trained using multi-modal loss that can generate multiple paths which accurately predict movement of NBA players. In addition, we showed that per-player fine-tuning can capture interesting and specific behavior of different players. The proposed approach outperformed state-of-the-art by a large margin, both in terms of standard prediction metrics and velocity error that better captures trajectory realism. As future work, we are exploring ideas to model the multi-modal behavior of the entire team, as well as opponents strategies that can counter such trajectories.

\clearpage

{%\small
\bibliographystyle{ieee_fullname}
\bibliography{bibliography}
}

\end{document}